\documentclass[journal]{IEEEtran}
\usepackage{amsmath,amsfonts}
\usepackage{algorithmic}
\usepackage{algorithm}
\usepackage{array}
\usepackage[caption=false,font=normalsize,labelfont=sf,textfont=sf]{subfig}
\usepackage{textcomp}
\usepackage{stfloats}
\usepackage{url}
\usepackage{verbatim}
\usepackage{graphicx}

\usepackage{cite}

\usepackage[colorlinks=true, linkcolor=blue, citecolor=blue, urlcolor=black]{hyperref}
\usepackage{graphicx}
\usepackage{booktabs}
\usepackage{pifont}
\usepackage{orcidlink}
\usepackage{multirow}
\usepackage{makecell}
\usepackage{xcolor}

\usepackage{cleveref}
\crefname{section}{Section}{Sections}
\crefname{figure}{Fig.}{Figs.}
\crefname{table}{Table}{Tables}
\begin{document}

\title{Argus: Leveraging Multiview Images for Improved 3-D Scene Understanding With Large \\ Language Models}

\author{Yifan Xu, Chao Zhang, Hanqi Jiang, Xiaoyan Wang, Ruifei Ma, Yiwei Li, Zihao Wu, Zeju Li, Xiangde Liu
        % <-this % stops a space
\thanks{Received 8 August 2024; revised 23 March 2025; accepted 12 June 2025. (\textit{corresponding author: Chao Zhang}).}
\thanks{Yifan Xu is with School of Computer Science and Engineering, Beihang University, Beijing 100191, China, also with Beijing Digital Native Digital City Research Center, Beijing 100084, China (email: \href{mailto:xudaxian2001@gmail.com}{xudaxian2001@gmail.com}).}
\thanks{Chao Zhang, Xiaoyan Wang, and Xiangde Liu are with Beijing Digital Native Digital City Research Center, Beijing 100084, China (email: \href{mailto:ariczhang2009@gmail.com}{ariczhang2009@gmail.com}, \href{mailto:wangxiaoyan@bdnrc.org.cn}{wangxiaoyan@bdnrc.org.cn}, \href{mailto:liuxiangde@bdnrc.org.cn}{liuxiangde@bdnrc.org.cn}).}
\thanks{Hanqi Jiang, Yiwei Li, Zihao Wu are with School of Computing, The University of Georgia, Athens, GA 30602-7404, USA (email: \href{mailto:hj67104@uga.edu}{hj67104@uga.edu}, \href{mailto:Yiwei.Li@uga.edu}{Yiwei.Li@uga.edu}, \href{mailto:Zihao.Wu1@uga.edu}{Zihao.Wu1@uga.edu}).}
\thanks{Ruifei Ma is with School of Computer and Communication Engineering, University of Science and Technology Beijing, Beijing 100083, China (email: \href{mailto:m202220831@xs.ustb.edu.cn}{m202220831@xs.ustb.edu.cn}).}
\thanks{Zeju Li is with the Department of Computer Science and Engineering, The Chinese University of Hong Kong, Hong Kong, China (email: \href{mailto:lizeju@link.cuhk.edu.hk}{lizeju@link.cuhk.edu.hk}).}
\thanks{Digital Object Identifier 10.1109/TNNLS.2025.3581411}
}

% The paper headers
\markboth{IEEE TRANSACTIONS ON NEURAL NETWORKS AND LEARNING SYSTEMS}{Shell \MakeLowercase{\textit{XU et al.}}: ARGUS: LEVERAGING MULTIVIEW IMAGES FOR IMPROVED 3-D SCENE UNDERSTANDING WITH LLMs}

%\IEEEpubid{0000--0000/00\$00.00~\copyright~2021 IEEE}
% Remember, if you use this you must call \IEEEpubidadjcol in the second
% column for its text to clear the IEEEpubid mark.

\maketitle

\begin{abstract}
  Advancements in foundation models have made it possible to conduct applications in various downstream tasks.
  Especially, the new era has witnessed a remarkable capability to extend Large Language Models (LLMs) for tackling tasks of 3D scene understanding. 
  Current methods rely heavily on 3D point clouds, but the 3D point cloud reconstruction of an indoor scene often results in information loss.
  Some textureless planes or repetitive patterns are prone to omission and manifest as voids within the reconstructed 3D point clouds.
  Besides, objects with complex structures tend to introduce distortion of details caused by misalignments between the captured images and the dense reconstructed point clouds.
  2D multi-view images present visual consistency with 3D point clouds and provide more detailed representations of scene components, which can naturally compensate for these deficiencies.
  Based on these insights, we propose \textit{Argus}, a novel 3D multimodal framework that leverages multi-view images for enhanced 3D scene understanding with LLMs.
  In general, Argus can be treated as a 3D Large Multimodal Foundation Model (3D-LMM) since it takes various modalities as input(text instructions, 2D multi-view images, and 3D point clouds) and expands the capability of LLMs to tackle 3D tasks.
  Argus involves fusing and integrating multi-view images and camera poses into view-as-scene features, which interact with the 3D features to create comprehensive and detailed 3D-aware scene embeddings.
  Our approach compensates for the information loss while reconstructing 3D point clouds and helps LLMs better understand the 3D world.
  Extensive experiments demonstrate that our method outperforms existing 3D-LMMs in various downstream tasks.
\end{abstract}


\section{Introduction}
\IEEEPARstart{F}{oundation} models have unlocked exciting possibilities for tackling various downstream tasks.
Among them, models like CLIP \cite{radford2021learning} and Uni3D \cite{zhou2023uni3d} explore alignment methods for multimodal data.
However, in light of the significant advantages of Large Language Models (LLMs) \cite{touvron2023llama}, \cite{touvron2023llama2}, \cite{vicuna2023}, \cite{chung2022scaling}, \cite{hua2023improving}, \cite{zhou2021topicbert} which focus on processing natural languages, researchers are exploring ways to employ vanilla LLMs to address a wide range of multimodal tasks \cite{alayrac2022flamingo}, \cite{liu2023llava}, \cite{gong2023multimodal}, \cite{chen2023octavius}.
2D vision-language tasks \cite{goyal2017making}, \cite{zhang2016yin}, \cite{chen2015microsoft} receive adequate attention, which try to equip LLMs with the ability to comprehend 2D images.
Mainstream techniques of 2D Vision Language Models (VLMs) \cite{alayrac2022flamingo}, \cite{liu2023llava}, \cite{kosmos-2}, \cite{li2023blip2}, \cite{instructblip}, \cite{liu2023improvedllava} can be divided into two paradigms.
One stream attempts to train the model with a large amount of text-image instruction-tuning data, as demonstrated by CLIP \cite{radford2021learning}, \cite{chen2023protoclip}, \cite{wang2023actionclip} and LLaVA \cite{liu2023llava}, \cite{liu2023improvedllava}.
The other paradigm concentrates on connecting pre-trained visual encoders and LLMs with additional learnable modules like perceiver in Flamingo \cite{alayrac2022flamingo} and Q-Former in BLIP-2 \cite{li2023blip2}, which align the latent space of vision embedding with text embedding of LLMs.

\begin{figure*}[tb]
    \centering
    \includegraphics[width=1.0\textwidth]{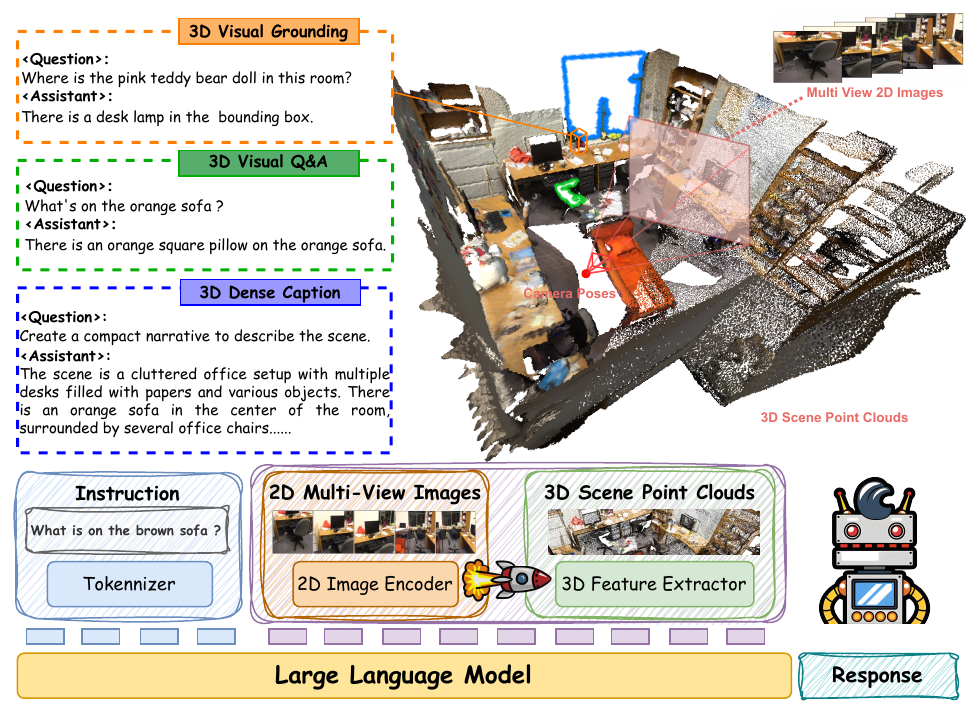}
    \caption{\textbf{High-level overview of our proposed method Argus}. It receives 2D multi-view images and 3D point cloud input and generates responses to text instructions. Argus can generalize to various downstream tasks in 3D scenes.}
    \label{fig-examples}
\end{figure*}

Similar to 2D vision-language, researchers have introduced LLMs for understanding 3D world \cite{guo2023point}, \cite{xu2023pointllm}, \cite{3dllm}, \cite{yin2023lamm}, \cite{li20243dmit}, \cite{chen2023ll3da}, \cite{huang2023chat}, \cite{chandhok2024scenegpt}.
Existing methods aim to align the 3D modality with the text modality to build 3D Large Multimodal Foundation Models (3D-LMMs).
This alignment enables LLMs to comprehend the 3D visual representations, serving as a premise for addressing general tasks in 3D scenes.
Integrating 3D representations into LLMs typically requires adding specialized structures that can effectively process spatial information.
3D-LLM \cite{3dllm} pioneered the injection of 3D spatial information into LLMs, marking a significant advancement in this area.
LL3DA \cite{chen2023ll3da} introduced visual prompts on the basis of the Q-Former architecture, opening up opportunities for LLMs to solve tasks in the 3D world.
By applying the 3D prompt, 3DMIT \cite{li20243dmit} eliminates an alignment stage between 3D scenes and natural language, but involves a fine-tuning stage of LLM.

Despite numerous studies on the 3D-LMMs topic, LLM-based methods for 3D scene understanding are still not fully mature.
3D point clouds depict the scene from a holistic perspective, providing structural and spatial information of the 3D world.
As a result, current methods \cite{3dllm}, \cite{yin2023lamm}, \cite{li20243dmit}, \cite{chen2023ll3da} rely heavily on 3D point clouds for scene understanding.
However, the process of reconstructing 3D point clouds of an indoor scene often results in information loss.
One issue is that regions characterized by textureless planes or repetitive patterns (such as uniform walls, homogeneous floor tiles, and glass surfaces) lack distinctive features within the captured images.
This deficiency significantly hampers the efficacy of stereo-matching algorithms, which rely on robust correspondence establishment between image pairs.
Consequently, these homogeneous areas are prone to omission and manifest as voids within the reconstructed 3D point clouds (see the blue area of the window in \cref{fig-examples}).
Another problem is that while dealing with objects with complex structures, misalignments between the captured images and the dense reconstructed point clouds have the propensity to introduce distortion of details.
Such distortion can prevent the full retention of detailed information, encompassing surface textures, boundary delineations, and micro characteristics.
These elements, often pivotal for comprehensive scene representation, may consequently appear blurred or indiscernible within the reconstructed 3D point clouds (see the green area of the chair in \cref{fig-examples}).
The degradation of the overall quality and utility of the reconstructed 3D point clouds undermines their capacity to faithfully represent the 3D scene, potentially leading to significant performance deterioration of existing methods.

2D multi-view images possess the capability to encapsulate virtually every region of a given scene, offering a detailed depiction of elements that significantly enhance the richness of 3D visual representations.
These details serve as compensation for the information loss in the 3D point cloud reconstruction process, thus deepening 3D-LMMs' comprehension of the complex 3D world.
Moreover, the inherent visual consistency between 2D multi-view images and 3D point clouds can establish a foundational bridge for this compensation without encountering semantic contradictions, facilitating a refined comprehension of the 3D scene.
It is worth noting that some LLM-based methods \cite{li20243dmit}, \cite{huang2023chat}, for understanding the 3D world, still require fine-tuning the LLM backbone, introducing a consumption of both training time and memory.
However, it is possible to bypass fine-tuning the LLM backbone through the utilization of a well-designed alignment module, such as Q-Former \cite{li2023blip2}, \cite{instructblip}.

Based on the aforementioned concerns, we propose Argus, a novel and flexible multimodal framework that leverages multi-view images for improved 3D scene understanding with LLMs.
Argus is a 3D-LMM since it takes various modalities as input and expands the capability of LLMs to 3D comprehension and reasoning.
Argus fuses and integrates 2D multi-view images and camera poses into view-as-scene features, which interact with the 3D features to derive comprehensive and detailed 3D scene representations, compensating for the information loss during point cloud reconstruction and helping LLMs better comprehend the 3D world.
Specifically, in addition to the LLM backbone, Argus consists of a novelly designed multi-view image fusion module and a 3D-aware Q-Former, which is an adaptation of the Q-Former architecture \cite{li2023blip2}.
For 2D multi-view images, we employ a Q-Former to obtain multi-view visual representations related to the text instructions.
All multi-view representations are processed by our fusion module, which generates view-as-scene features.
These features encapsulate detailed 3D scene information, stemming from the integration of multi-view images and their corresponding camera poses.
Then, we introduce the 3D-aware Q-Former, which is a specialized module designed to facilitate the interaction between 3D features and view-as-scene features, generating comprehensive 3D-aware embeddings that conform to the directivity of the text instructions from multi-view images and 3D point clouds.
The output of 3D-aware Q-Former serves as the scene-level representations, strongly correlated with both 3D point clouds and multi-view images. 
The fused scene-level representations are projected to the embedding space of LLM to generate responses.
Through this approach, we obtain the detailed information provided by 2D multi-view images, which improves the 3D scene understanding of LLMs via 3D-aware Q-Former.
With the assistance of 2D multi-view images, our framework can effectively tackle 3D scene understanding tasks.
This alignment of modalities, executed with precision and care, emerges as a foundation for enriching 3D representations and unlocking the capabilities of our framework towards a new horizon of effectiveness.
It is important to note the flexibility of our framework, as both the fusion module and the 3D-aware Q-Former can independently receive their respective modalities of data, outperforming 2D-only or 3D-only methods.
In terms of the fusion module, the view-as-scene features are directly sent into the LLM for downstream tasks.
Alternatively, the 3D-aware Q-Former receives only 3D features as input and outputs 3D-aware embeddings into LLM.
Furthermore, due to the advantages of the Q-Former architecture, the fine-tuning stage of the LLM backbone is unnecessary.

Our key contributions can be summarized as follows:
\begin{itemize}
    \item We propose our 3D large multimodal foundation model named Argus, which is an innovative and flexible framework designed for enhancing the comprehension of the 3D world by utilizing 2D multi-view images and 3D point clouds effectively.
    \item We investigate an effective approach to fuse and integrate multi-view images and camera poses into view-as-scene features by our designed fusion module, which interact with the 3D features to create comprehensive and detailed 3D scene representations by our 3D-aware Q-Former. In this way, the information loss during 3D point cloud reconstruction can be compensated, contributing to better 3D scene understanding with LLMs.
    \item Extensive experiments show that Argus achieves superior performance on downstream 3D scene understanding tasks compared to other 3D-LMMs. We also demonstrate that our designed fusion module and 3D-aware Q-Former can mutually enhance the performance.
\end{itemize}

\section{Related Work}

\subsection{3D-Language Scene Understanding}
3D-Language Scene Understanding is a burgeoning field that explores the intersection of 3D scene vision understanding \cite{chandhok2024scenegpt}, \cite{ding2022language}, \cite{zhu2023vista} and natural language processing.
Research in this area encompasses a variety of tasks, including 3D Question Answering (3D-QA) \cite{azuma_2022_CVPR}, \cite{ma2022sqa3d}, \cite{wang2022visrecall}, \cite{zhao2022towards}, 3D Visual Grounding (3D-VG) \cite{achlioptas2020referit_3d}, \cite{chen2020scanrefer}, \cite{wu2022eda}, and 3D captioning \cite{chen2021scan2cap}, among others. 
3D-QA requires a model to provide natural language answers to questions related to the given 3D scenes. 
This task requires the model to comprehend the scenes and generate accurate responses. 
3D-VG necessitates a model to interpret natural language queries and identify specific instances within the 3D scene, localizing the target object and providing its coordinates. 
This task demands a strong understanding of objects' spatial relationships. 
3D captioning evaluates a model's capability to generate captions that precisely describe the 3D scenes. 
This task requires the model to understand 3D scenes holistically and depict them in coherent and descriptive language.
Sophisticated models designed for 3D-language tasks often leverage related large-scale annotated datasets, such as ScanNet \cite{dai2017scannet}, 3RScan \cite{Wald2019RIO}, and Matterport3D \cite{Matterport3D}, to learn rich representations of 3D scenes and texts. 
However, the scarcity of 3D-language datasets presents a significant challenge in training well-generalized models for 3D scene understanding. 
In this paper, we propose Argus, which addresses the issue by leveraging obtainable 2D multi-view images in real-world scenarios, to enhance 3D-LMMs' performance for 3D scene understanding.

\subsection{Multimodal Large Language Models}
Due to the exceeding advantages of LLMs in language tasks, researchers have increasingly sought to expand the capability of LLMs to multimodal tasks \cite{xu2023multimodal}, \cite{yin2023survey}. 
Specifically, some methods \cite{alayrac2022flamingo}, \cite{liu2023llava}, \cite{kosmos-2}, \cite{li2023blip2}, \cite{instructblip}, \cite{liu2023improvedllava}, \cite{zhang2023llamaadapter}, \cite{gao2023llamaadapterv2}, \cite{chen2023shikra} concentrate on integrating image modalities into LLMs. 
LLaVA \cite{liu2023llava}, \cite{liu2023improvedllava} and LAMM \cite{yin2023lamm} have employed instruction tuning with image-text pairs to enhance the multimodal performance of MLLMs. 
Approaches like Q-Former \cite{li2023blip2} and Perceiver \cite{alayrac2022flamingo} aim to align the vision embedding spaces with the text embedding space of LLMs, thus avoiding fine-tuning the parameters of LLMs. 
InstructBlip \cite{instructblip} is a combination of instruction tuning and Q-Former, unlocking more possibilities for vision-language understanding.

3D-LLM \cite{3dllm} first bridges LLMs with 3D modality and stands out as a pioneering endeavor, marking the inaugural fusion of the 3D world with LLMs.
Additionally, LL3DA \cite{chen2023ll3da} enhances the capabilities by introducing visual prompts on the shoulder of Q-Former, thus facilitating deeper comprehension of 3D environments through interactive visual clicks and boxes. 
3DMIT \cite{li20243dmit} proposes an efficient tuning method for 3D-LLMs, and eliminates the alignment stage between 3D scenes and language.
However, current methods mainly consider 3D representations of scenes, which tend to provide a comprehensive perspective of the entire scene but neglect the detailed information provided by multi-view images.
In this paper, we present our 3D-LMM, Argus, which involves the fusion of multi-view images and camera poses into view-as-scene features for improving the understanding of the 3D world.
Distinguishing it from 3D-LLM, which reconstructed 3D features based on multi-view images, our method complements 3D features with the fused multi-view representation by interacting with Q-Former's 3D learnable queries.

\begin{figure*}[tb]
    \centering
    \includegraphics[width=1.0\textwidth]{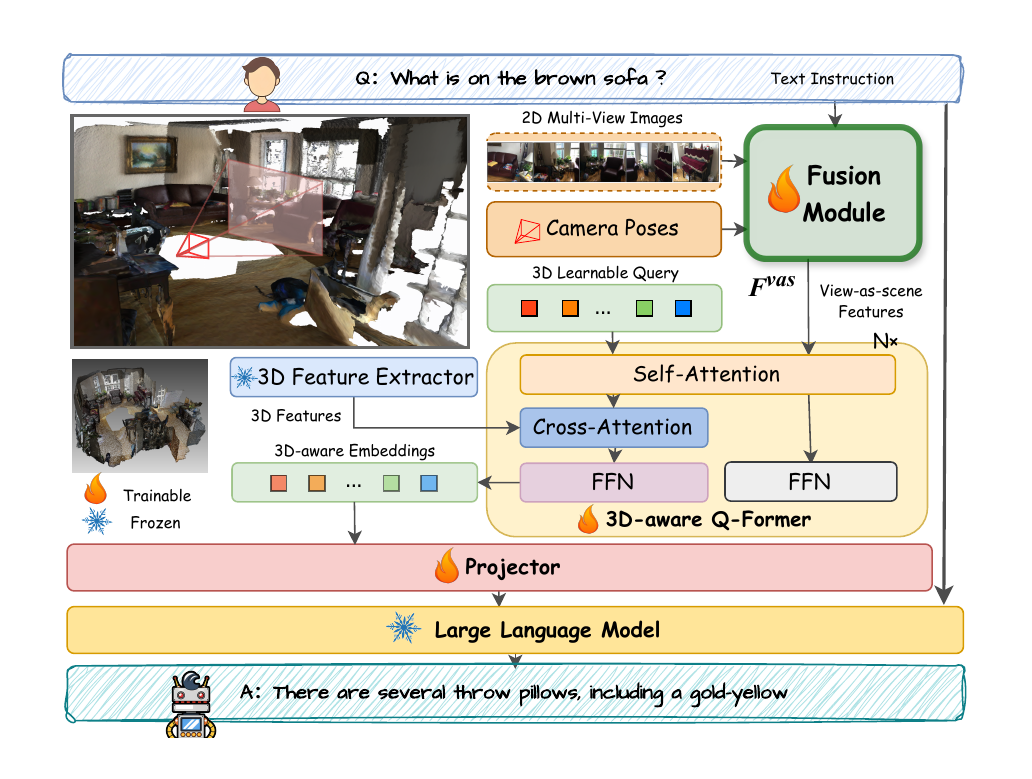}
    \caption{\textbf{Overall pipeline of our method.}
    The fusion module aggregates multi-view images and their corresponding camera poses into view-as-scene features.
    The learnable queries of 3D-aware Q-Former interact with the 3D features and view-as-scene features. The outputs of 3D-aware Q-Former(3D-aware embeddings) are then projected to the embedding space of LLM.}
    \label{fig-method}
\end{figure*}

\section{Methodology}

As illustrated in \cref{fig-method}, we propose Argus, which utilizes multi-view images for improved 3D scene understanding with LLMs.
Our framework consists of three fundamental components: a fusion module to aggregate view-as-scene features, a 3D-aware Q-Former, and a frozen LLM.
Multi-view images of a specific scene, along with their corresponding camera poses, are first integrated into our designed fusion module, which is composed of a 2D Q-Former \cite{li2023blip2}, a two-layer MLP, and four stacked transformer \cite{vaswani2017attention} layers.
This fusion module derives scene-level representations, which are denoted as view-as-scene features, and provides detailed information about the 3D scene that may be lost during 3D point cloud reconstruction.
We extract 3D features of the scene using our 3D feature extractor, such as EPCL \cite{huangepcl}.
We introduce the 3D Learnable Query, which is a set of trainable vectors that serve as a bridge between the 3D features extracted from point clouds and the view-as-scene features derived from multi-view images.
Our designed 3D-aware Q-Former enables these learnable queries to interact with the 3D features and view-as-scene features and learn unified features that conform to the directivity of the text instructions from multi-view images and 3D point clouds, and generates 3D-aware embeddings that incorporate comprehensive and detailed information from 3D point clouds, multi-view images, and textual instructions.
These 3D-aware embeddings are then projected into the input embedding space of LLM, which generates answers to the input text instructions.

\subsection{Revisiting Q-Former}\label{Sec3.1}

Q-Former \cite{li2023blip2}, \cite{instructblip} is proposed to bootstrap vision-language pre-training with frozen image encoders and LLMs. 
Q-Former consists of an image transformer and a text transformer with shared weights of self-attention layers.
The image transformer interacts with the frozen image encoder to extract visual representations, while the text transformer plays the role of an encoder and a decoder simultaneously.

During training, Q-Former constructs learnable queries as the input of the image transformer, which fuse text embedding through the shared self-attention layers while acquiring image information through the cross-attention layers.
Q-Former bridges the gap between vision and language, without tuning the image encoder and the large language model.
During inference, only visual features are processed by Q-Former.

Our framework borrows some ideas from Q-Former. 
In our framework, the fusion module employs a 2D Q-Former to extract multi-view features.
We also apply our 3D-aware Q-Former, which enables the learnable queries to interact with the 3D features and view-as-scene features and outputs 3D-aware embeddings to LLM.

\subsection{Pre-trained 2D Image Encoder and 3D Feature Extractor} 

Now we have a few multi-view images for a specific 3D scene. 
Following BLIP-2, we adopt pre-trained ViT-g/14 from EVA-CLIP \cite{fang2023eva} as our image encoder to extract image features with dimension $256\times1408$.
We explore two different 3D feature extractors in our framework.
One is EPCL \cite{huangepcl}, which applies pre-trained CLIP to extract 3D point cloud features with a fixed dimension $256\times1024$ in an effective and efficient way.
Another one is directly reconstructing $N\times D$-dim 3D feature from multi-view images, following the step of 3D-LLM \cite{3dllm}, where $N$ and $D$ refer to the point cloud number and feature dimension.

Both the extracted 2D multi-view image features and 3D features are aligned with text modality, establishing a solid foundation prepared for further 3D scene understanding.

\begin{figure*}[tb]
    \centering
    \includegraphics[width=1.0\textwidth]{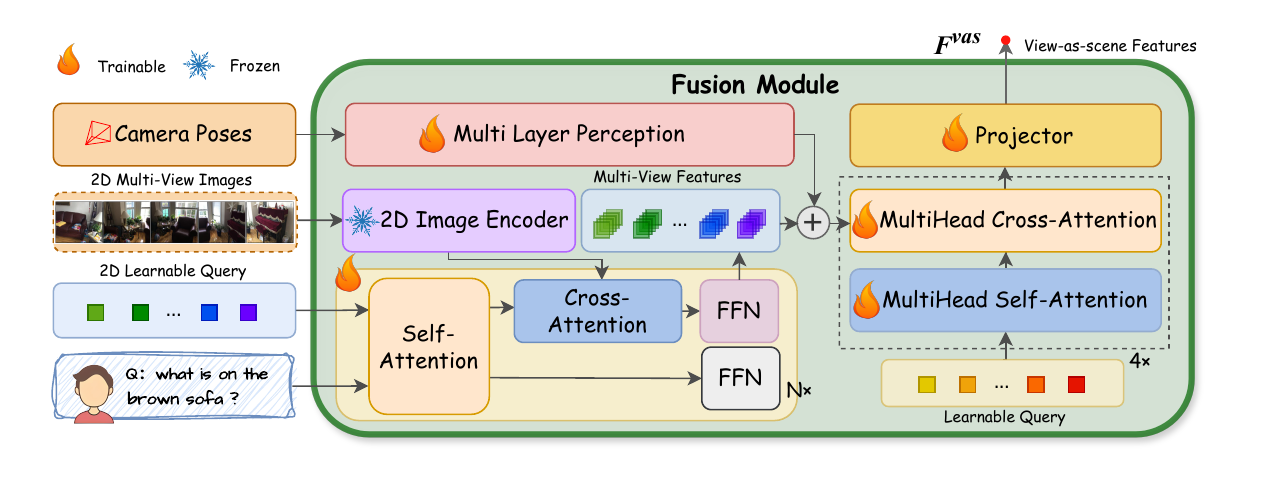}
    \caption{\textbf{Structure of the fusion module.}
    The module involves two key steps: extracting multi-view features with spatial information and aggregating these features as view-as-scene features.
    By capturing detailed information about the scene, the view-as-scene features facilitate the acquisition of comprehensive 3D-aware embeddings.}
    \label{fig-fusion}
\end{figure*}

\subsection{Multi-view Images Fusion}\label{vas}

Unlike 3D point clouds, which tend to depict the holistic scene, 2D multi-view images can provide more detailed information about the scene components.
This discrepancy presents an opportunity to enrich 3D visual representations and deepen 3D-LMMs' comprehension of the complex 3D world.
Specifically, as shown in \cref{fig-fusion}, we design a novel fusion module for aggregating the multi-view images for subsequent enrichment, which involves extracting multi-view features with spatial information and then aggregating these features to derive view-as-scene features.

\textbf{Step 1: Extracting multi-view features with spatial information.}
All multi-view images $\{I_i\}_{i=1}^n$ of a specific 3D scene are processed by the pre-trained image encoder ViT to obtain image features $\{f_i\}_{i=1}^n$, with $n$ as the number of images.
For each image $I_i$, we get its visual feature $f_i$ as follows:
\begin{equation}
    f_i = \text{ViT}({I_i}).
\end{equation}

Then we apply a 2D Q-Former, which is revisited in \cref{Sec3.1}, to derive multi-view features $\{q_i\}_{i=1}^n$ related to the text instructions.
Taking feature $f_i$ as an example:
\begin{equation}
    q_i = \text{Q-Former}(f_i).
\end{equation}

However, features in $\{q_i\}_{i=1}^n$ still lack of spatial information.
Inherently, each multi-view image possesses a camera pose, which refers to the view's position and orientation relative to the 3D scene, thus reflecting some spatial information indirectly.
Consequently, we utilize a Flatten layer followed by a two-layer MLP to transform the camera poses $\{p_i\}_{i=1}^n$ of multi-view images into position embeddings $\{PosEmb_i\}_{i=1}^n$:
\begin{equation}\label{pos_emb}
    PosEmb_i = \text{MLP}(\text{Flatten}(p_i)).
\end{equation}
Through this transformation, we encode spatial information directly into the feature representations.
These position embeddings are then added to the multi-view image features, ensuring that the spatial context of each image is retained in the final feature representation $F_i^{\text{view}}$:
\begin{equation}\label{f_view}
    F_i^{\text{view}} = q_i + PosEmb_i.
\end{equation}

\textbf{Step2: Aggregating multi-view features as view-as-scene features.} 
As each multi-view feature in $\{F^{\text{view}}_i\}_{i=1}^n$ merely represents partial information of the scene, we explore fusing the multi-view features as view-as-scene features using four stacked transformer layers.
Specifically, we employ a multi-head cross-attention (MHCA) mechanism \cite{vaswani2017attention} after performing self-attention, enabling the module to perform attentive interactions between the multi-view features and a set of trainable queries $Q$.
The number of queries is set to 32, which is generally acknowledged in Q-Former architecture, and it impacts the richness of representative patterns and computational overhead during training.
The aggregation process can be described as follows:
\begin{equation}
    F^{\text{vas}} = \text{Proj}(\text{MHCA}(q=Q,kv=\{F^{\text{view}}_i\}_{i=1}^n)),
\end{equation}
where $q$, $k$, and $v$ denote query, key, and value in the attention mechanism, and Proj($\cdot$) represents a simple linear layer.
This mechanism allows the model to perform attentive interactions between the features of different views, effectively integrating spatial context information from multiple perspectives.

Our fusion module integrates spatial information into multi-view features and fuses these features as view-as-scene features, providing detailed information about the scene.

\subsection{3D-Aware Q-Former}

Considering the constructed 3D features of point clouds may face information loss and the view-as-scene features derived from \cref{vas} contain detailed information, we design a 3D-aware Q-Former to extract both comprehensive and detailed cues through interactions between the 3D features and view-as-scene features.
% Given the 3D features and view-as-scene features from \cref{vas}, we design a 3D-aware Q-Former to perform interactions between the 3D features and view-as-scene features.

Specifically, the 3D-aware Q-Former incorporates a set of trainable vectors, 3D Learnable Query, which engage in a two-step interaction process.
First, these queries interact with the view-as-scene features through a weight-shared self-attention layer.
The self-attention mechanism ensures that the queries are informed by the rich and detailed visual context.
This is crucial for resolving ambiguities and compensating for missing information caused by misalignments.
It effectively helps the queries capture detailed information from the multi-view visual representation that might otherwise be lost during point cloud construction.
Subsequently, the processed queries interact with the input 3D features via a cross-attention layer.
The cross-attention mechanism enables the queries to incorporate the holistic information of the scene.
Through the combined action of both the self-attention and cross-attention mechanisms, the 3D-aware Q-Former empowers the learnable queries to learn unified representations that conform to the directivity of the text instructions from multi-view images and 3D point clouds, which capture both comprehensive and detailed information of 3D scenes during training.
The comprehensive information allows the model to understand the scene from a holistic perspective, while the detailed information helps to supplement ambiguities and information loss.
The 3D-aware embeddings output by the 3D-aware Q-Former are then projected into the input embedding space of the LLM using a linear layer.

The 3D-aware Q-Former effectively bridges the gap between 3D modality and text modality.
This is achieved by exploiting the visual consistency between 3D features and multi-view features, which are aligned with the text modality by the fusion module.
Furthermore, the interactions within the 3D-aware Q-Former help derive meaningful 3D-aware embeddings, which encapsulate comprehensive and detailed information about the scene.
The 3D-aware embeddings enhance the LLM's ability to conduct accurate and contextually relevant comprehension of 3D scenes.

\subsection{Training Pipeline}\label{trainingpipeline}

We adopt a three-stage training approach for our framework to enhance its performance.
(1) In the first stage, we pre-train the 3D-aware Q-Former using extensive data, encompassing tasks such as 3D-QA, scene captioning, embodied dialogue, and planning.
(2) In the second stage, we pre-train both the fusion module and the 3D-aware Q-Former using the same dataset.
These two pre-training stages equip the LLM in our framework with a strong understanding of 3D concepts, enabling the LLM to proficiently handle a wide array of tasks.
(3) In the third stage, we fine-tune the fusion module and the 3D-aware Q-Former using task-specific data, further enhancing its capability to handle each specific task.

Throughout all training stages, our framework, shown in \cref{fig-method}, remains unchanged.
During the first stage, the fusion module still provides view-as-scene features, but its parameters, initialized from the LAVIS library \cite{li2022lavis}, are frozen.

Moreover, we employ the standard cross-entropy loss to model language discrepancy, which helps the model learn to generate contextually appropriate and coherent responses.
It is worth noting that we keep all parameters of the LLM backbone frozen during both the training and fine-tuning stages.

\section{Experiments}

\subsection{Training Data}

This section provides detailed statistics on the datasets used to train our model.
Our 3D scenes are sourced from ScanNet \cite{dai2017scannet}, a comprehensive 3D indoor dataset with 1513 diverse 3D indoor scenes that encompass various environments such as apartments, living rooms, kitchens, and bedrooms.
In the pre-training stage, our language annotations are obtained from 3DMIT \cite{li20243dmit} and a subset of ScanNet provided by 3D-LLM \cite{3dllm}.
For 3DMIT, we collect approximately 25k VQA pairs.
Regarding the ScanNet subset from 3D-LLM, we utilize all annotations, including about 1k scene descriptions, 15k question-answering pairs, 17k lines of embodied planning, and 8k lines of multi-turn embodied dialogues.

\subsection{Implementation Details}

We adopt ViT-g/14 from EVA-CLIP \cite{fang2023eva} as our 2D image encoder.
% explore
We employ two 3D feature extractors, EPCL \cite{huangepcl} and the method for reconstructing 3D features by 3D-LLM \cite{3dllm}.
For each specific scene, we choose 100 multi-view images, which will be discussed in \cref{view_number}.
We initialize the Q-Former in the fusion module and the 3D-aware Q-Former from checkpoints released in the LAVIS library \cite{li2022lavis}.
We apply two LLM backbones with different architectures: FlanT5-XL \cite{chung2022scaling} with an encoder-decoder architecture and Vicuna7B \cite{vicuna2023} with a decoder-only architecture.
During pre-training in stages 1 and 2, we use the AdamW \cite{loshchilov2017decoupled} optimizer with $\beta_1=0.9, \beta_2=0.999$ and a weight decay of 0.05.
The learning rate warms up from $10^{-8}$ and increases linearly to $10^{-4}$, then gradually decays to $10^{-5}$ using a cosine decay.
During fine-tuning, we used the AdamW optimizer similarly, only adjusting the learning rate decay from $10^{-5}$ to $10^{-6}$.
Setting the batch size to 64, our pre-training process takes approximately 10 hours on eight A800-80G GPUs, with each fine-tuning task requiring about 5 hours.

\begin{table*}
\centering
\caption{Experimental results of 3D-QA on ScanQA validation and test dataset.}
\setlength{\tabcolsep}{5pt}
\label{ScanQA}
\begin{tabular}{c|cccccc|cccccc} 
\toprule
\multirow{2}{*}{Method} & \multicolumn{6}{c|}{Validation}                                                               & \multicolumn{6}{c}{Test w/ object}                                                             \\
                        & EM            & Bleu-1        & Bleu-4        & Meteor        & Rouge-L       & CIDEr         & EM            & Bleu-1        & Bleu-4        & Meteor        & Rouge-L       & CIDEr          \\ 
\hline
\multicolumn{13}{l}{\textit{\textbf{Expert Models}}}                                                                                                                                                                     \\ 
\hline
VoteNet \cite{ding2019votenet}+MCAN \cite{yu2019deep}           & 17.3          & 28.0          & 6.2           & 11.4          & 29.8          & 54.7          & 19.7          & 29.5          & 6.0           & 12.0          & 30.9          & 58.2           \\
ScanRefer \cite{chen2020scanrefer}+MCAN \cite{yu2019deep}         & 18.6          & 26.9          & 7.9           & 11.5          & 30.0          & 55.4          & 20.6          & 27.9          & 7.5           & 11.9          & 30.7          & 57.4           \\
ScanQA \cite{azuma_2022_CVPR}                 & 21.0          & 30.2          & 10.1          & 13.1          & 33.3          & 64.9          & 23.5          & 31.6          & 12.0          & 13.5          & 34.3          & 67.3           \\ 
\hline
\multicolumn{13}{l}{\textit{\textbf{General Multimodal LLMs}}}                                                                                                                                                          \\ 
\hline
3D-LLM (Flamingo-3B) \cite{3dllm}    & 20.4          & 30.3          & 7.2           & 12.2          & 32.3          & 59.2          & 23.2          & 32.6          & 8.4           & 13.5          & 34.8          & 65.6           \\
3D-LLM (Opt-1.3B) \cite{3dllm}       & 19.3          & \underline{35.9}          & 9.4           & 13.8          & 34.0          & 63.8          & 19.1          & 37.3          & 10.7          & 14.3          & 34.5          & 67.1           \\
3D-LLM (FlanT5-3B) \cite{3dllm} & 20.5          & \textbf{39.3} & 12.0          & 14.5          & 35.7          & 69.4          & 19.1          & \underline{38.3}          & 11.6          & 14.9          & 35.3          & 69.6           \\
3DMIT (Vicuna-7B) \cite{li20243dmit}       & 13.0          & 27.6          & 5.2           & 10.7          & 26.2          & 48.0          & -             & -             & -             & -             & -             & -              \\
NaviLLM(Vicuna-7B) \cite{zheng2023learning}     & \textbf{23.0} & -             & 12.5          & 15.4          & \textbf{38.4} & 75.9          & \textbf{26.3} & -             & \underline{13.9}          & \textbf{16.6} & \textbf{40.2} & \textbf{80.8}  \\
LL3DA (Opt-1.3B-w/oVP) \cite{chen2023ll3da}   & -             & -             & 11.9          & 14.0          & 33.9          & 67.9          & -             & -             & -             & -             & -             & -              \\
LL3DA (Opt-1.3B-w/VP) \cite{chen2023ll3da}   & -             & -             & \textbf{13.5} & \underline{15.8}          & 37.3          & \underline{76.8}          & -             & -             & \textbf{16.4} & 14.0          & 38.2          & 78.2           \\ 
\hline
Argus (FlanT5-3B)         & \underline{21.8}          & \textbf{39.3} & \underline{13.0}          & \textbf{15.9} & \underline{38.2}          & \textbf{76.9} & \underline{25.9}          & \textbf{39.3} & 11.1          & \underline{16.4}          & \underline{40.1}          & \underline{79.3}           \\
\bottomrule
\end{tabular}
\flushleft{"VP" refers to visual prompts in LL3DA. Bold and underlined numbers are the top two scores.}
\end{table*}

\subsection{Evaluation Results on 3D-QA}

3D Question Answering (3D-QA) tasks involve generating responses to natural language queries about a 3D scene.
To prepare our model, we fine-tune it on the ScanQA \cite{azuma_2022_CVPR} dataset, which consists of over 40,000 3D-QA pairs derived from 800 indoor scenes within the ScanNet \cite{dai2017scannet} dataset.
Subsequently, we evaluate our model's performance using the ScanQA validation dataset.

\textbf{Baselines.}
We categorize our compared baselines into two main groups: Expert Models and General Multimodal LLMs.
(1) Expert Models:
\textbf{VoteNet+MCAN}: This method combines VoteNet \cite{ding2019votenet} for object detection with MCAN \cite{yu2019deep}, a standard VQA model, to generate answers to 3D questions.
\textbf{ScanRefer+MCAN}: This method integrates ScanRefer \cite{chen2020scanrefer} for object localization with MCAN \cite{yu2019deep} to provide accurate answers to 3D questions.
\textbf{ScanQA}: This benchmark method uses VoteNet for object proposal generation and fuses the proposals with language embeddings to generate high-quality answers \cite{azuma_2022_CVPR}.
(2) General Multimodal LLMs:
\textbf{3D-LLM}: This family of models aligns 3D scenes with text features using 2D Vision-and-Language Models as a backbone and projects 3D features into the input embedding space of VLM \cite{3dllm}.
\textbf{3DMIT}: This method utilizes an efficient 3D multimodal instructions tuning approach to train LLMs, enhancing their ability to understand and generate responses for 3D scenes \cite{li20243dmit}.
\textbf{NaviLLM}: This model uses a simple transformer to combine observations and historical information, creating a complete scene representation for navigation tasks \cite{zheng2023learning}.
\textbf{LL3DA}: This model combines text instructions and visual interactions, extracting features that enhance its ability to follow instructions accurately and generate contextually relevant responses \cite{chen2023ll3da}.
We present all the results obtained through task-specific fine-tuning.

\textbf{Evaluation Metrics.}
For the 3D-QA task, we report rich metrics, including EM, BLEU-1, BLEU-2, BLEU-3, BLEU-4, Meteor, Rouge-L, and CIDEr, to evaluate the answers generated by LLM.
Among these metrics, EM calculates the exact match scores with the ground truth, while BLEU-n represents the answer's coarse-grained similarity.
Rouge-L tends to consider the integrity and order of generated answers with the expected output, while Meteor and CIDEr focus more on semantic consistency.

\textbf{Performance Analysis.}
As illustrated in \cref{ScanQA}, our approach achieves competitive performance in 3D-QA across all metrics, demonstrating its ability to generate both exact and diverse responses.
We observe a significant improvement of Argus over the expert method ScanQA \cite{azuma_2022_CVPR} and methods that explicitly introduce object features such as ScanRefer \cite{chen2020scanrefer}+MCAN \cite{yu2019deep}, VoteNet \cite{ding2019votenet}+MCAN \cite{yu2019deep}, and 3DMIT \cite{li20243dmit}.
This indicates that Argus could conduct visual reasoning about objects and their spatial relationships, which is attributed to the details brought by multi-view images.

Our method demonstrates significant improvements over the 3D-LLM family \cite{3dllm} across most metrics, with notable enhancements in EM, Meteor, Rouge-L, and CIDEr scores.
Specifically, on the ScanQA validation set, while utilizing the same LLM backbone and similar Q-Former architecture, we observe a +7.5 improvement in CIDEr and a +2.5 increase in Rouge-L.
While our method achieves superior performance in many metrics, it is important to note that the 3D-LLM family shows slightly higher BLEU-1 and BLEU-4 scores in some cases.
This indicates that while our approach excels in generating contextually rich and diverse responses, there is still room for improvement in terms of n-gram precision for longer sequences.
% Our method remarkably surpasses the 3D-LLM family across all metrics.
We outperform NaviLLM \cite{zheng2023learning} in Bleu-4, Meteor, and CIDEr on the ScanQA validation set, while achieving neck-and-neck performance on other metrics.
% However, NaviLLM utilizes large quantities of extra training data, such as R2R, SOON, REVERIE, CVDN, and LLaVA-23k.
% We prove that our method is on par with NaviLLM while utilizing significantly less training data (only ScanNet).
We find that the performance gain in LL3DA \cite{chen2023ll3da} mainly comes from introducing visual prompts, which refer to user clicks and 3D box annotations. 
Without explicitly introducing visual prompts, our method outperforms LL3DA by around +9 CIDEr scores.
We also surpass LL3DA with visual prompts across all metrics except Bleu-4.
On the ScanQA test set with objects, our method achieves comparable performance with NaviLLM and substantially outperforms all other state-of-the-art approaches.

In summary, these results highlight the enhanced capability of our method to extract comprehensive and detailed 3D-aware embeddings by leveraging multi-view images, which leads to improved performance in the 3D-QA task.

\begin{table}
\centering
\caption{Comparison of Training Requirements between Argus and other General Multimodal LLMs.}
\label{TrainingReq}
\setlength{\tabcolsep}{3pt}
\begin{tabular}{c|c|c|c|c} 
\toprule
\multirow{2}{*}{Method} & \multicolumn{4}{c}{Training Requirement}                                                                                                                                                                                                         \\ 
\cline{2-5}
                        & Dataset                                                              & Annotation                                                                             & Pair  & Resource                                                                 \\ 
\hline
3D-LLM \cite{3dllm}                 & \begin{tabular}[c]{@{}c@{}}ScanNet, HM3D\\and Objaverse\end{tabular} & GPT                                                                                    & 525k+ & \begin{tabular}[c]{@{}c@{}}8*A100\\\textgreater{}48h\end{tabular}        \\ 
\hline
3DMIT \cite{li20243dmit}                  & ScanNet                                                              & ScanQA, GPT                                                                            & 75k   & \begin{tabular}[c]{@{}c@{}}8*A100\\\textasciitilde{}24h\end{tabular}     \\ 
\hline
NaviLLM \cite{zheng2023learning}                & \begin{tabular}[c]{@{}c@{}}Matterport3D\\and ScanNet\end{tabular}    & \begin{tabular}[c]{@{}c@{}}LLaVA-23k, R2R,\\CVDN, SOON,\\~REVERIE, ScanQA\end{tabular} & 80k   & \begin{tabular}[c]{@{}c@{}}8*A100\\\textasciitilde{}48h\end{tabular}     \\ 
\hline
LL3DA \cite{chen2023ll3da}                  & ScanNet                                                              & \begin{tabular}[c]{@{}c@{}}ScanRefer, Nr3D,\\~ ScanQA, 3D-LLM\end{tabular}             & 120k  & \begin{tabular}[c]{@{}c@{}}8*3090\\\textasciitilde{}24h\end{tabular}  \\ 
\hline
Argus                   & ScanNet                                                              & 3D-LLM, 3DMIT                                                                          & 75k   & \begin{tabular}[c]{@{}c@{}}8*A800\\\textasciitilde{}15h\end{tabular}     \\
\bottomrule
\end{tabular}
\flushleft{"GPT" represents annotations generated by GPT. "3D-LLM" refers to the ScanNet subset of 3D-LLM, and "3DMIT" denotes the QA part of 3DMIT.}
\end{table}

\textbf{Training Resource Analysis.}
\cref{TrainingReq} shows the training requirements of Argus and the compared general multimodal LLMs.
3D-LLM \cite{3dllm} leverages substantial training datasets, including 475k language annotations from Objaverse and 50k from ScanNet, and some from HM3D.
3DMIT \cite{li20243dmit} incorporates 75k language annotations from ScanNet \cite{dai2017scannet}, encompassing tasks such as VQA, grounding, captioning, conversation, and multi-choice.
LL3DA \cite{chen2023ll3da} pre-trains its model using ScanRefer \cite{chen2020scanrefer}, Nr3D \cite{achlioptas2020referit_3d}, ScanQA \cite{azuma_2022_CVPR}, and the ScanNet subset released by 3D-LLM, totaling 120k annotations.
It is important to note that while NaviLLM \cite{zheng2023learning} utilizes a similar scale of training data in terms of the number of annotations, it incorporates additional navigation-related datasets from scenes in Matterport3D \cite{Matterport3D}, such as CVDN \cite{thomason2020vision}, SOON \cite{Zhu_2021_CVPR}, R2R \cite{mattersim}, and REVERIE \cite{reverie}.
These datasets provide NaviLLM with a broader range of navigation scenarios, which may contribute to its performance in the 3D-QA task.
% NaviLLM\cite{zheng2023learning} trains on a combined dataset from CVDN\cite{thomason2020vision}, SOON\cite{Zhu_2021_CVPR}, R2R\cite{mattersim}, REVERIE\cite{reverie}, ScanQA, and LLaVA-23k\cite{liu2023llava}, totaling 80k annotations on different scene datasets.
Notably, without an explicit alignment stage, both 3DMIT and NaviLLM fine-tune the LLM backbones.
In contrast, our method, which leverages multi-view images to enhance 3D scene understanding, achieves competitive performance only on ScanNet with a total of 75k language annotations.
Moreover, we do not fine-tune the LLM backbone, saving significant computational costs.

\subsection{Evaluation Results on 3D-VG}

3D-VG requires a model to comprehend natural language queries and accurately identify specific instances within a 3D scene, localizing the target object and providing its coordinates. 
As 3D-VG demands detailed object information, we consider using PointGroup \cite{jiang2020pointgroup} to segment objects in the scene and extract their features by Uni3D \cite{zhou2023uni3d}.
The object features are projected into the input embedding space of the LLM and concatenated with the 3D-aware embeddings, which are output by the 3D-aware Q-Former.
We expect the LLM to generate the object identifier and then create the corresponding 3D bounding box.
We pre-train our model with FlanT5 backbone and 3D features from 3D-LLM \cite{3dllm}.
Subsequently, we fine-tune our pre-trained model on ScanRefer before conducting evaluations.

\textbf{Baselines.}
(1) Expert Models:
ScanRefer \cite{chen2020scanrefer} utilizes a pre-trained VoteNet backbone alongside a trained GRU to choose a matching bounding box.
(2) General Multimodal LLMs:
LLM-Grounder \cite{yang2023llm} achieves zero-shot 3D visual grounding using an open-vocabulary grounding tool, with GPT-4 as the assistant.
3D-LLM \cite{3dllm} predicts bounding boxes as location tokens added to the language model's vocabularies and learned from scratch.
3DMIT \cite{li20243dmit} predicts the object ID and its bounding box using the center's coordinates, length, width, and height.
We present all the results obtained through task-specific fine-tuning.

\textbf{Evaluation Metrics.}
For the 3D-VG task, we evaluate the model's performance in 3D grounding using the metrics Acc@0.25 and Acc@0.5.
These metrics measure the accuracy of bounding box predictions by comparing their Intersection-over-Union (IoU) with the ground-truth box, with IoU thresholds of 0.25 and 0.5, respectively.

\begin{table}[tbp!]
  \caption{Experimental results of 3D-VG on ScanRefer.}
  \label{grounding}
  \centering
  \setlength{\tabcolsep}{2mm}
  \begin{tabular}{@{}c|cc@{}}
    \toprule
    Methods & Acc@0.25 & Acc@0.5 \\
    \hline
    ScanRefer \cite{chen2020scanrefer}  & \textbf{37.3} & \underline{24.3}  \\
    \hline
    LLM-Grounder (GPT-4) \cite{yang2023llm} & 17.1 & 5.3  \\
    3D-LLM (Flamingo-3B) \cite{3dllm}  & 21.2 & -  \\
    3D-LLM (FlanT5-3B) \cite{3dllm} & 30.3 & -  \\
    3DMIT (Vicuna-7B) \cite{li20243dmit} & 10.2 & 7.1  \\
    % Chat3D-v2 & 35.9 & 30.4  \\
    \hline
    Argus (Scene) & 4.1 & 4.0  \\
    Argus (Objects) & 25.4 & 21.8  \\
    Argus (Scene+Objects) & \underline{35.2} & \textbf{29.4}  \\
    % Ours(Vicuna) & 36.4 & 30.2  \\
  \bottomrule
  \end{tabular}
\end{table}

\textbf{Result Analysis.}
As shown in \cref{grounding}, when only provided with 3D-aware embeddings, the LLM fails to accurately identify objects.
This limitation stems from 3D-aware embeddings describing the scene holistically, with implicit object spatial relationships.
By incorporating object features alongside 3D-aware embeddings, our method achieves a significant improvement on the 3D-VG task.
This highlights that our 3D-aware embeddings provide LLM with rich information besides object features during visual grounding.
Via the improvement brought by multi-view images, our method outperforms general multimodal LLMs and achieves competitive performance compared to expert methods.

\begin{table*}[tbp!]
    \caption{Evaluations on embodied dialogue, scene description, and embodied planning.}
    \label{tasks}
    \centering
    \begin{tabular}{c|c|cccccccc}
        \toprule
        Task & Method & EM & Bleu-1 & Bleu-2 & Bleu-3 & Bleu-4 & Meteor & Rouge-L & CIDEr \\ 
        \midrule
        \multirow{4}*{Embodied Dialogue}  & Opt-1.3B (zero-shot) & - & 2.4 & 1.1 & 0.5 & 0.2 & 5.6 & 4.8 & 0.3 \\ 
        ~ & LLAMA-7B (zero-shot) & - & 4.1 & 1.8 & 0.9 & 0.5 & 7.8 & 6.7 & 0.3 \\ 
        ~ & LL3DA (Opt-1.3B) & - & \underline{41.4} & \textbf{32.6} & \underline{27.5} & \textbf{24.0} & \textbf{23.5} & \underline{40.6} & \underline{190.0} \\ 
        ~ & Argus (FlanT5-3B) & 6.6 & \textbf{41.8} & \underline{32.5} & \textbf{27.6} & \underline{23.5} & \underline{20.3} & \textbf{41.5} & \textbf{197.1} \\ 
        \midrule
        \multirow{4}*{Scene Description} & Opt-1.3B (zero-shot) & - & 15.8 & 6.1 & 2.1 & 0.8 & 8.4 & 11.7 & 0.0 \\ 
        ~ & LLAMA-7B (zero-shot) & - & 19.3 & 7.7 & 2.8 & 0.9 & 7.0 & 12.3 & 0.2 \\ 
        ~ & LL3DA (Opt-1.3B) & - & \textbf{43.0} & \textbf{26.7} & \textbf{16.0} & \textbf{9.0} & \underline{14.7} & \underline{24.8} & \underline{1.0} \\ 
        ~ & Argus (FlanT5-3B) & 0.1 & \underline{34.0} & \underline{21.0} & \underline{13.5} & \underline{8.8} & \textbf{15.2} & \textbf{29.2} & \textbf{1.4} \\ 
        \midrule
        \multirow{4}*{Embodied Planning} & Opt-1.3B (zero-shot) & - & 1.3 & 0.6 & 0.3 & 0.1 & 0.2 & 3.6 & 0.2 \\ 
        ~ & LLAMA-7B (zero-shot) & - & 2.2 & 1.1 & 0.6 & 0.3 & 3.5 & 4.7 & 0.1 \\
        ~ & LL3DA (Opt-1.3B) & - & \textbf{40.7} & \textbf{27.2} & \textbf{18.6} & \underline{13.0} & \underline{17.1} & \underline{39.3} & \underline{128.8} \\ 
        ~ & Argus (FlanT5-3B) & 5.7 & \underline{35.1} & \underline{24.9} & \underline{17.4} & \textbf{13.1} & \textbf{18.5} & \textbf{41.2} & \textbf{135.2} \\ 
        \bottomrule
    \end{tabular}
\end{table*}

\begin{table*}
  \caption{Evaluation of the benefit from multi-view images.}
  \label{2D3D}
  \centering
  \begin{tabular}{@{}c|c|cccccccc@{}}
    \toprule
    Methods & Modality & EM & Bleu-1 & Bleu-2 & Bleu-3 & Bleu-4 & Meteor & Rouge-L & CIDEr\\
    \midrule
    SingleImage (Flamingo-3B) & 2D & \textbf{16.9} & 23.8 & 14.5 & 9.2 & \textbf{8.5} & 10.7 & 29.6 & 52.0 \\
    MultiView (Flamingo-3B) & 2D & 18.8 & 25.6 & 15.2 & 9.2 & 8.4 & 11.3 & 31.1 & 55 \\
    SingleImage (FlanT5-3B) & 2D & 13.3 & 28.6 & 15.1 & 9.0 & 5.1 & 10.6 & 25.8 & 42.6 \\
    MultiView (FlanT5-3B) & 2D & 13.6 & 29.7 & 16.2 & 9.8 & 5.9 & 11.3 & 26.6 & 45.7 \\
    Argus (FlanT5-3B) & 2D & \textbf{16.9} & \textbf{35.0} & \textbf{20.4} & \textbf{12.3} & 7.1 & \textbf{12.9} & \textbf{33.2} & \textbf{59.9} \\
    \midrule
    3D-LLM (FlanT5-3B) & 3D & 20.5 & \textbf{39.3} & \textbf{25.2} & \textbf{18.4} & 12.0 & 14.5 & 35.7 & 69.4 \\
    Argus (FlanT5-3B) & 3D & \textbf{21.1} & 37.0 & 24.0 & 17.0 & \textbf{12.1} & \textbf{14.9} & \textbf{36.7} & \textbf{72.5} \\
    \midrule
    Argus (FlanT5-3B) & 2D + 3D & 21.8 & 39.3 & 25.3 & 18.7 & 13.0 & 15.9 & 38.2 & 76.9 \\
  \bottomrule
  \end{tabular}
\end{table*}

\begin{figure}
    \centering
    \includegraphics[width=1.0\linewidth]{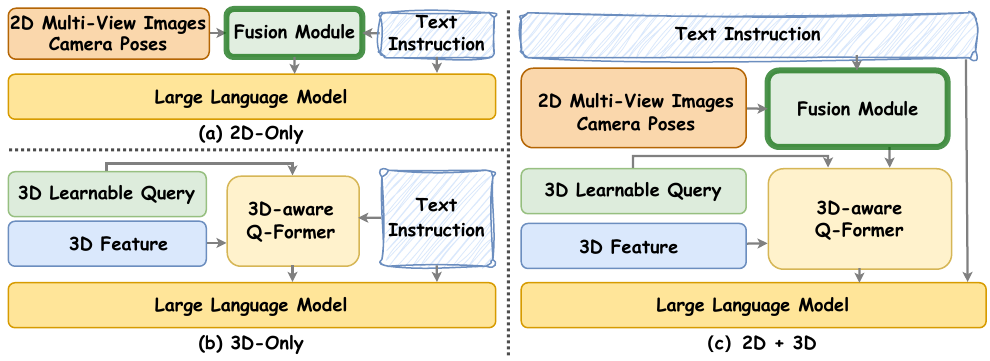}
    \caption{Three inference structures of Argus. After training with 2D multi-view images, Argus can receive the input of (a) only 2D images, (b) only 3D point clouds, and (c) the combination of both  2D images and 3D point clouds.}
    \label{fig:inferencemode}
\end{figure}

\subsection{Evaluations on More Tasks}

As shown in \cref{tasks}, we carry out extensive evaluations on three more tasks: embodied dialogue, scene description, and embodied planning. 
Following LL3DA \cite{chen2023ll3da}, we set the scenes with IDs less than 600 as the training set, and the rest as the validation set.
We evaluate the generated natural language responses of our method and compare zero-shot LLMs (Opt-1.3B and LLAMA-7B) and LL3DA under the EM, Bleu, Meteor, Rouge-L, and CIDEr metrics. 
From the table, we can see that our method significantly outperforms zero-shot LLMs.
LL3DA's superior performance on BLEU metrics can be attributed to its explicit use of visual prompts, which guide the model to focus on the objects in clicks or 3D boxes, resulting in more precise local matches with ground truth answers.
Our method, which leverages multi-view images to supplement the information loss during the 3D scene construction, provides a more comprehensive understanding of the scene, leading to more diverse and contextually rich responses, as evidenced by higher Meteor, Rouge-L, and CIDEr scores.

\subsection{Benefit From Multi-view Images}

After training with 2D multi-view images, Argus can receive the input of 2D images, 3D point clouds, or their combination.
We separate the fusion module and the 3D-aware Q-Former from the fine-tuned model to evaluate their individual performances.
As depicted in \cref{fig:inferencemode} (a), Argus can process only 2D multi-view images as input to the fusion module and generate view-as-scene features, which are directly fed into the LLM backbone for downstream tasks.
As shown in \cref{fig:inferencemode} (b), Argus can take only 3D point clouds (3D features) as input, which the 3D-aware Q-Former processes, and outputs 3D-aware embeddings to the LLM.
Notably, in our experiments, SingleImage trains the models by replacing the 3D inputs of 3D-LLMs with single-image features.
MultiView, on the other hand, replaces the 3D inputs of 3D-LLMs with concatenated multi-view image features.
These results are based on the 3D-LLM proposed in \cite{3dllm}.

It is important to note that when 2D multi-view images are not integrated, our method simplifies to 3D-LLM.
As shown in \cref{2D3D}, when the fusion module and 3D-aware Q-Former are used together (\cref{fig:inferencemode} (c)), our method consistently outperforms 3D-LLM, except for an equal score in the Bleu-1 metric.
This shows that using 2D multi-view images can effectively improve 3D scene understanding.
Furthermore, our fusion module outperforms both the SingleImage and MultiView methods.
The performance improvement is substantial, with a CIDEr metric score that is more than 5 points higher than the best 2D-only method.
Additionally, when the 3D-aware Q-Former receives only 3D point clouds as input, it outperforms 3D-LLM in terms of EM, Meteor, Rouge-L, and CIDEr.
From the experimental results, we can conclude that in addition to the effectiveness of 2D multi-view images, both the fusion module and the 3D-aware Q-Former can independently process their respective data modalities.
They outperform methods that use only 2D or 3D data, indicating a mutually beneficial performance enhancement.
Therefore, whether using only 2D images or 3D scene data, our approach consistently improves performance.

\subsection{Ablation Study}

We perform four ablation studies to assess the impact of selecting diverse quantities of multi-view images per scene, the effectiveness of incorporating different 3D features, employing diverse training stages, and utilizing various fusion module designs.
For simplicity, we report only the experimental results obtained with the FlanT5 backbone and 3D-LLM feature on the ScanQA validation dataset while conducting training on the ScanQA training dataset.

\begin{table}[tbp!]
  \caption{Effect of the number of multi-view images per scene.}
  \label{viewnumber}
  \centering
  \setlength{\tabcolsep}{1mm}
  \begin{tabular}{@{}c|cccccc@{}}
    \toprule
    \makecell{Quantity of \\ Multi-view Images} & EM & Bleu-1 & Bleu-4 & Meteor & Rouge-L & CIDEr\\
    \midrule
    80 / each scene & 20.9 & 37.8 & 12.0 & 15.4 & 37.1 & 73.7 \\
    100 / each scene & 21.4 & 38.1 & 12.5 & 15.3 & 37.5 & 74.5 \\
    120 / each scene & 21.1 & 37.6 & 11.8 & 15.1 & 37.6 & 74.1 \\
  \bottomrule
  \end{tabular}
\end{table}

\begin{table}[tbp!]
  \caption{Effectiveness of our fusion module design.}
  \label{fusionmodule}
  \centering
  \setlength{\tabcolsep}{1.5mm}
  \begin{tabular}{@{}c|cccccc@{}}
    \toprule
    Methods & EM & Bleu-1 & Bleu-4 & Meteor & Rouge-L & CIDEr\\
    \midrule
    Average & 18.3 & 35.4 & 10.7 & 13.6 & 34.7 & 69.7 \\
    Max & 19.9 & 37.5 & 10.8 & 14.4 & 35.3 & 72.2 \\
    \midrule
    Ours  & \textbf{21.4} & \textbf{38.1} & \textbf{12.5} & \textbf{15.3} & \textbf{37.5} & \textbf{74.5} \\
  \bottomrule
  \end{tabular}
\end{table}

\begin{table}[tbp!]
  \caption{Effect of applying different training stages.}
  \label{stages}
  \centering
  \setlength{\tabcolsep}{1mm}
  \begin{tabular}{@{}ccc|cccccccc@{}}
    \toprule
    Stage1 & Stage2 & Stage3 & EM & Bleu-1 & Bleu-4 & Meteor & Rouge-L & CIDEr\\
    \midrule
    \ding{55} & \ding{55} & \ding{51} & 21.4 & 38.1 & 12.5 & 15.3 & 37.5 & 74.5 \\
    \ding{51} & \ding{55} & \ding{51} & 21.1 & 37.9 & 12.5 & 15.4 & 37.8 & 75.0 \\
    \ding{55} & \ding{51} & \ding{51} & 21.6 & 39.0 & 12.9 & 15.8 & 38.1 & 76.5 \\
    \ding{51} & \ding{51} & \ding{51} & 21.8 & 39.3 & 13.0 & 15.9 & 38.2 & 76.9 \\
  \bottomrule
  \end{tabular}
\end{table}

\textbf{Effect of selecting a diverse quantity of multi-view images for each scene.} \label{view_number}
To enhance 3D scene understanding using multi-view images, the selection of the quantity of images is critical.
In our evaluation, we consider selecting 80, 100, and 120 multi-view images per scene.
We observe a marginal performance decrease when choosing either 80 or 120 images per scene, as presented in \cref{viewnumber}.
Insufficient images fail to capture detailed information, while an excess of images can negatively impact 3D vision representation.
Based on these observations, we selected 100 multi-view images for each scene in all our experiments.

\textbf{Effectiveness of our fusion module design.}
After obtaining multi-view image features with spatial information, we explore different methods to create a comprehensive view-as-scene feature.
"Average" refers to applying the average operation of all multi-view features in the view-number dimension, while "Max" involves selecting the maximum value of the same dimension.
As \cref{fusionmodule} demonstrates, there are significant differences in outcomes when using either the maximum feature or the average of all features.
This discrepancy underscores that averaging may discard crucial information, while taking the maximum preserves the most critical feature.
Nonetheless, our approach, which utilizes a four-layer stacked transformer, achieves the best performance, showcasing our ability to derive the optimal view-as-scene feature.

\begin{table}[tbp!]
  \caption{Effect of different 3D features.}
  \label{differentfeature}
  \centering
  \setlength{\tabcolsep}{1pt}
  \begin{tabular}{@{}c|c|cccccccc@{}}
    \toprule
    Methods & Feature & EM & Bleu-1 & Bleu-4 & Meteor & Rouge-L & CIDEr\\
    \midrule
    3D-LLM (FlanT5-3B) & 3D-LLM & 20.5 & 39.3 & 12.0 & 14.5 & 35.7 & 69.4 \\
    Argus (FlanT5-3B)  & 3D-LLM & 21.8 & 39.3 & 13.0 & 15.9 & 38.2 & 76.9 \\
    Argus (FlanT5-3B)  & EPCL & 21.5 & 38.8 & 12.1 & 15.6 & 37.7 & 75.1 \\
  \bottomrule
  \end{tabular}
\end{table}

\begin{table}[tbp!]
  \caption{Effectiveness of camera poses.}
  \label{camerapose}
  \centering
  \setlength{\tabcolsep}{1mm}
  \begin{tabular}{@{}c|cccccc@{}}
    \toprule
    Methods & EM & Bleu-1 & Bleu-4 & Meteor & Rouge-L & CIDEr\\
    \midrule
    w/o camera poses & 21.1 & 37.2 & 11.4 & 15.2 & 36.8 & 72.6 \\
    w/ camera poses  & 21.4 & 38.1 & 12.5 & 15.3 & 37.5 & 74.5 \\
  \bottomrule
  \end{tabular}
\end{table}

\begin{table}[tbp!]
  \caption{Effect of different 3D-aware Q-Former initialization weights.}
  \label{init}
  \centering
  \setlength{\tabcolsep}{2pt}
  \begin{tabular}{@{}c|c|cccccccc@{}}
    \toprule
    3D Features & \makecell{Initialization \\ Weight} & EM & Bleu-1 & Bleu-4 & Meteor & Rouge-L & CIDEr\\
    \midrule
    3D-LLM & Random & 20.3 & 37.6 & 11.8 & 15.0 & 36.9 & 71.2 \\
    3D-LLM & Argus (BLIP-2) & 21.4 & 38.1 & 12.5 & 15.3 & 37.5 & 74.5 \\
    \midrule
    EPCL & Random & 19.5 & 36.9 & 11.6 & 14.7 & 35.9 & 69.2 \\
    EPCL & Argus (BLIP-2) & 20.9 & 37.7 & 12.0 & 15.1 & 36.3 & 71.9 \\
  \bottomrule
  \end{tabular}
\end{table}

\begin{figure*}
    \centering
    \includegraphics[width=1.0\textwidth]{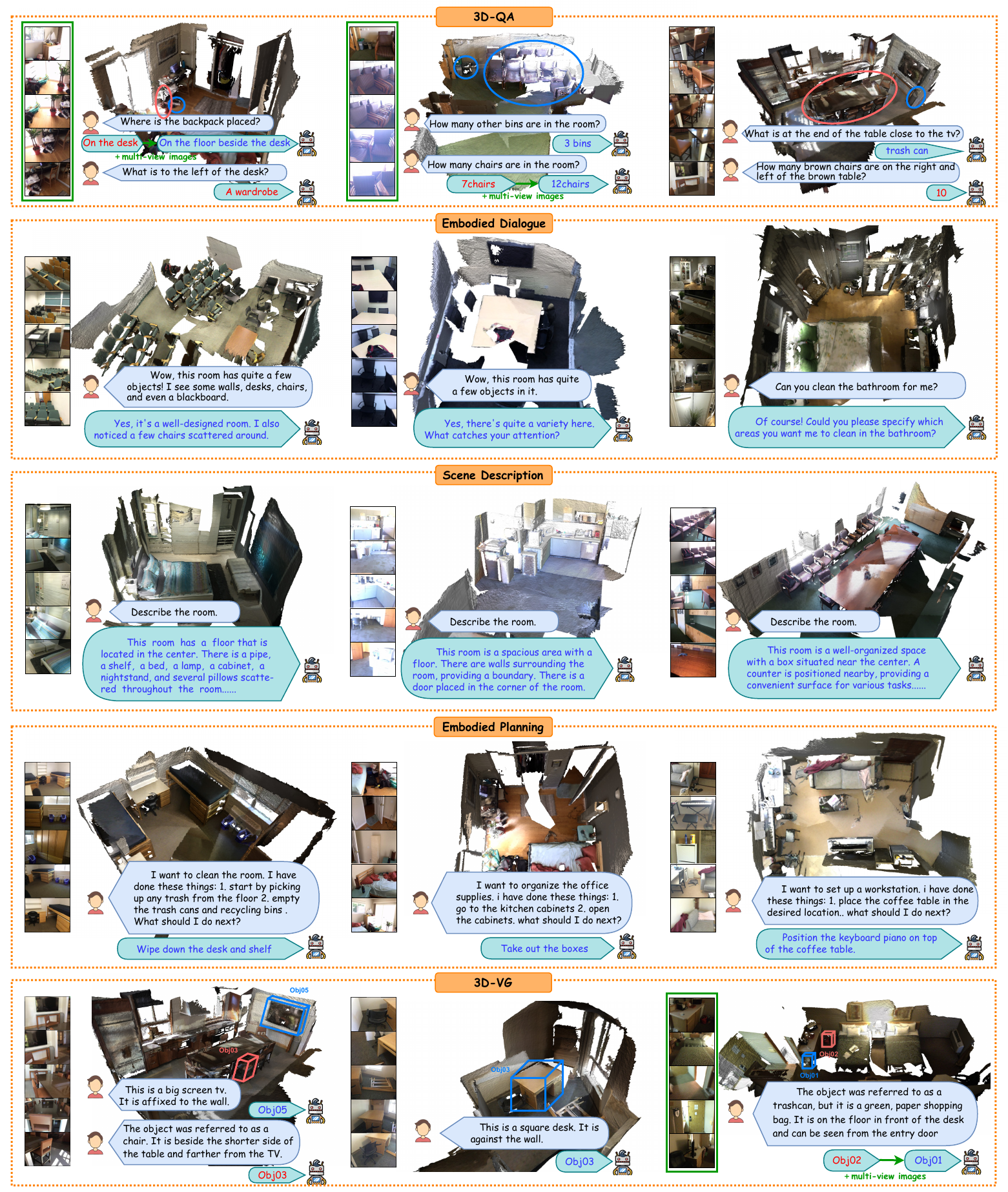}
    \caption{\textbf{Qualitative Results.} We provide several visualization results of various tasks in diverse 3D environments, including 3D-QA, embodied dialogue, scene description, embodied planing and 3D-VG.
    Failure cases are visualized in red. The green arrow denotes leveraging 2D multi-view images.}
    \label{examples}
\end{figure*}

\textbf{Effect of applying different training stages.}
We explore the impact of applying different training stages (See \cref{trainingpipeline}) on model performance.
We can infer from the results in \cref{stages} that if we directly apply stage 3, the performance experiences a slight decrease.
Due to the substantial learnable parameters, it is difficult to enable the 3D-aware Q-Former to fully perceive 3D information while solely training with task-specific data.
We also observe performance improvements when pre-training the 3D-aware Q-Former separately or along with the fusion module, which can be attributed to optimizing the network in the direction of perceiving 3D information.
While applying all training stages, we find that our method achieves the best performance, demonstrating the effectiveness of our training approach.

\textbf{Effect of different 3D features.}
We conduct an ablation study to discuss the effectiveness of different 3D features.
We mainly adopt two 3D features.
One is to reconstruct 3D features by 3D-LLM \cite{3dllm}, and the other is EPCL \cite{huangepcl}, which leverages a CLIP-based point cloud encoder to directly extract 3D point cloud features.
The experimental results in \cref{differentfeature} show that applying these two different 3D features can achieve close performance, demonstrating that our method is well-scalable and can adapt to diverse 3D features.
In some ways, this is also attributed to aggregating multi-view image features to obtain a better 3D vision representation.

It is worth noting that, regardless of whether EPCL or 3D-LLM features are applied, our method significantly outperforms 3D-LLM.

\textbf{Effectiveness of camera poses.}
We assess the effect of integrating camera poses as position embeddings into the multi-view features in the fusion module.
As demonstrated in \cref{camerapose}, our approach consistently outperforms the method that does not use camera poses across all metrics, especially achieving a +2 score improvement on CIDEr.
This indicates that incorporating camera poses as position embeddings enhances our view-as-scene feature.
It enables the view-as-scene feature to capture more spatial information, leading to more comprehensive 3D-aware embeddings when conducting interactions in 3D-aware Q-Former.

\textbf{Effect of different 3D-aware Q-Former initialization weights.}
We also evaluate the effectiveness of various initialization weights for the 3D-aware Q-Former.
Results presented in \cref{init} indicate that initializing the 3D-aware Q-Former with the Q-Former weights from BLIP-2 leads to improved performance when utilizing both 3D-LLM and EPCL features.
This suggests that our method can better grasp 3D concepts when initialized with weights based on 2D images, confirming the representative relationship between 2D multi-view images and 3D point clouds.
Moreover, this finding indirectly supports the feasibility of enhancing 3D scene understanding through the utilization of 2D multi-view images.

\subsection{Further Discussion}

\textbf{Dataset biases and generalization.}
ScanNet primarily consists of indoor scenes, which may introduce biases towards specific types of environments and objects commonly found in indoor settings.
This could limit the model's ability to generalize to outdoor scenes or other contexts where the visual and spatial characteristics differ significantly.
Moreover, the language annotations in ScanNet are based on human-provided descriptions and questions, which may introduce biases related to the specific language used, the level of detail, and the types of questions asked.
This could affect the model's ability to understand and generate responses for different linguistic styles or more complex queries.

Despite the indoor bias, the diversity of scenes within ScanNet (e.g., apartments, living rooms, kitchens, and bedrooms) helps Argus learn a broad range of indoor environments.
This diversity enables the model to generalize well across different types of indoor scenes, capturing the commonalities and variations within these environments.
However, although Argus may generalize to outdoor scenes with the assistance of LLMs' commonsense knowledge, incorporating additional training datasets that include outdoor scenes would help Argus learn the unique characteristics of outdoor environments, such as larger spatial scales and different lighting conditions.

\textbf{Visualization cases and limitations.}
In \cref{examples}, we showcase several visualization examples of diverse tasks generated by Argus. These tasks encompass 3D-QA, embodied dialogue, scene description, embodied planning, and 3D-VG.
By leveraging multi-view images for supplementation, Argus can effectively address the point cloud voids caused by information loss and exhibit robust performance in downstream tasks (e.g., cases 1 and 2 of 3D-QA).
Concurrently, with the help of multi-view images, Argus can generate precise responses to queries about small or detailed objects (e.g., case 3 of 3D-VG).
% Even though an irrelevant question is given, Argus can still generate a relatively reasonable response.

Nevertheless, the spatial information encapsulated in camera poses remains notably weak and inadequate.
As a result, Argus may encounter difficulties in handling the spatial location relationships of objects, particularly when explicit directions are provided (e.g., cases 1 and 3 of 3D-QA, and case 1 of 3D-VG).
Future development efforts will concentrate on strengthening the model's capacity to comprehend spatial relationships.

% The integration of multi-view images into 3D point clouds requires substantial computational resources.
% Processing large volumes of data, aligning images, and projecting them onto the 3D model can be time-consuming and resource-intensive, potentially limiting real-time applications or those requiring rapid updates.

\section{Conclusion}

In this paper, we propose a novel framework, Argus, that enhances 3D scene understanding through the utilization of multi-view images.
We investigate an effective approach to fuse and integrate multi-view images and their corresponding camera poses into view-as-scene features by our designed Q-Former, which interacts with the 3D features to create comprehensive and detailed 3D-aware embeddings by our 3D-aware Q-Former.
Extensive experiments demonstrate that our approach outperforms various 3D-LMMs in downstream tasks.
We believe that our method holds significant implications for future research by leveraging multi-view images for improved 3D scene understanding.
We hope that our approach will inspire further innovations in the design of methods and training strategies for 3D-LMMs.

% references section
\bibliographystyle{IEEEtran}
\bibliography{IEEEabrv,reference}

\end{document}